\documentclass{article}
\PassOptionsToPackage{numbers,compress}{natbib}
\usepackage[final]{neurips_2020}

\usepackage[utf8]{inputenc} % allow utf-8 input
\usepackage[T1]{fontenc}    % use 8-bit T1 fonts
\usepackage{hyperref}       % hyperlinks
\usepackage{url}            % simple URL typesetting
\usepackage{booktabs}       % professional-quality tables
\usepackage{amsfonts}       % blackboard math symbols
\usepackage{nicefrac}       % compact symbols for 1/2, etc.
\usepackage{microtype}      % microtypography

\usepackage{graphicx}
\usepackage{amsmath}
\usepackage{amssymb}
\usepackage{mhchem}
\usepackage{wrapfig}

\title{On the equivalence of molecular graph convolution
and molecular wave function with poor basis set}

\author{
Masashi Tsubaki\\
National Institute of Advanced Industrial Science and Technology\\
\texttt{tsubaki.masashi@aist.go.jp}\\
\AND
Teruyasu Mizoguchi\\
Institute of Industrial Science, University of Tokyo\\
\texttt{teru@iis.u-tokyo.ac.jp}
}

\begin{document}

\maketitle

\begin{abstract}
In this study, we demonstrate that
the linear combination of atomic orbitals (LCAO),
an approximation of quantum physics introduced by Pauling and Lennard-Jones in the 1920s,
corresponds to graph convolutional networks (GCNs) for molecules.
However, GCNs involve unnecessary nonlinearity and deep architecture.
We also verify that molecular GCNs are based on
a poor basis function set compared with the standard one
used in theoretical calculations or quantum chemical simulations.
From these observations, we describe the quantum deep field (QDF),
a machine learning (ML) model based on an underlying quantum physics,
in particular the density functional theory (DFT).
We believe that the QDF model can be easily understood
because it can be regarded as a single linear layer GCN.
Moreover, it uses two vanilla feedforward neural networks to learn
an energy functional and a Hohenberg--Kohn map that
have nonlinearities inherent in quantum physics and the DFT.
For molecular energy prediction tasks,
we demonstrated the viability of an ``extrapolation,''
in which we trained a QDF model with small molecules,
tested it with large molecules, and achieved high extrapolation performance.
This will lead to reliable and practical applications for discovering effective materials.
The implementation is available at
\url{https://github.com/masashitsubaki/QuantumDeepField_molecule}.
\end{abstract}

\section{Introduction}

Recently, graph convolutional networks (GCNs)~\cite{scarselli2008graph,
kipf2016semi} have been applied to molecular graphs.
Although numerous variants of the molecular GCN have been
developed~\cite{kearnes2016molecular,gilmer2017neural,chen2019graph}
(Section~2.1), they have a basic computational procedure:
the GCN model (1)~considers that each node (i.e., atom) of a molecular graph
has a multidimensional variable (i.e., the atom feature vector),
(2)~uses the convolutional operation to update the feature vectors
according to the graph structure defined by
the adjacency or distance matrix between the atoms in the molecule,
and finally (3)~outputs a value (e.g., the energy of the molecule)
via a readout function (e.g., the sum/mean of the updated vectors).
Deep neural networks (DNNs) are used in the convolutional operation.
Therefore, GCNs involve strong nonlinearity when modeling
the molecular graph structure and have achieved good prediction performance
on large-scale benchmark datasets, such as QM9~\cite{ramakrishnan2014quantum}.

In this study, from the perspective of quantum physics,
we demonstrate that extant molecular GCNs involve
unnecessary nonlinearity and deep architecture.
We first describe an approximation of quantum physics
introduced by Pauling and Lennard-Jones in
the 1920s~\cite{pauling1960nature,lennard1929electronic,mcquarrie1997physical},
which states that the superposition of atomic wave functions
(called orbitals) is based on their linear combination (Section~2.2).
We demonstrate that this linear superposition/combination corresponds to
the convolutional operation in GCNs; that is, the nonlinear DNN used in
the above (2) is not required for modeling the molecular structure (Section~3).
Additionally, in the linear superposition/combination,  % theないの？
although the number of wave functions/orbitals (or basis functions)
and the types of each basis function are important,
the molecular GCNs do not consider these points.
In particular, the reason for performance degradation in deeper GCNs
has been discussed recently in~\cite{kipf2016semi,xu2019powerful};
however, it is trivial with regard to molecules.
The molecular GCNs are built on a poor and incorrect basis function set compared with
the standard one used in theoretical calculations or quantum chemical simulations.

\begin{figure}[t]
\begin{center}
\includegraphics[width=14cm, bb=0 350 1920 1000]{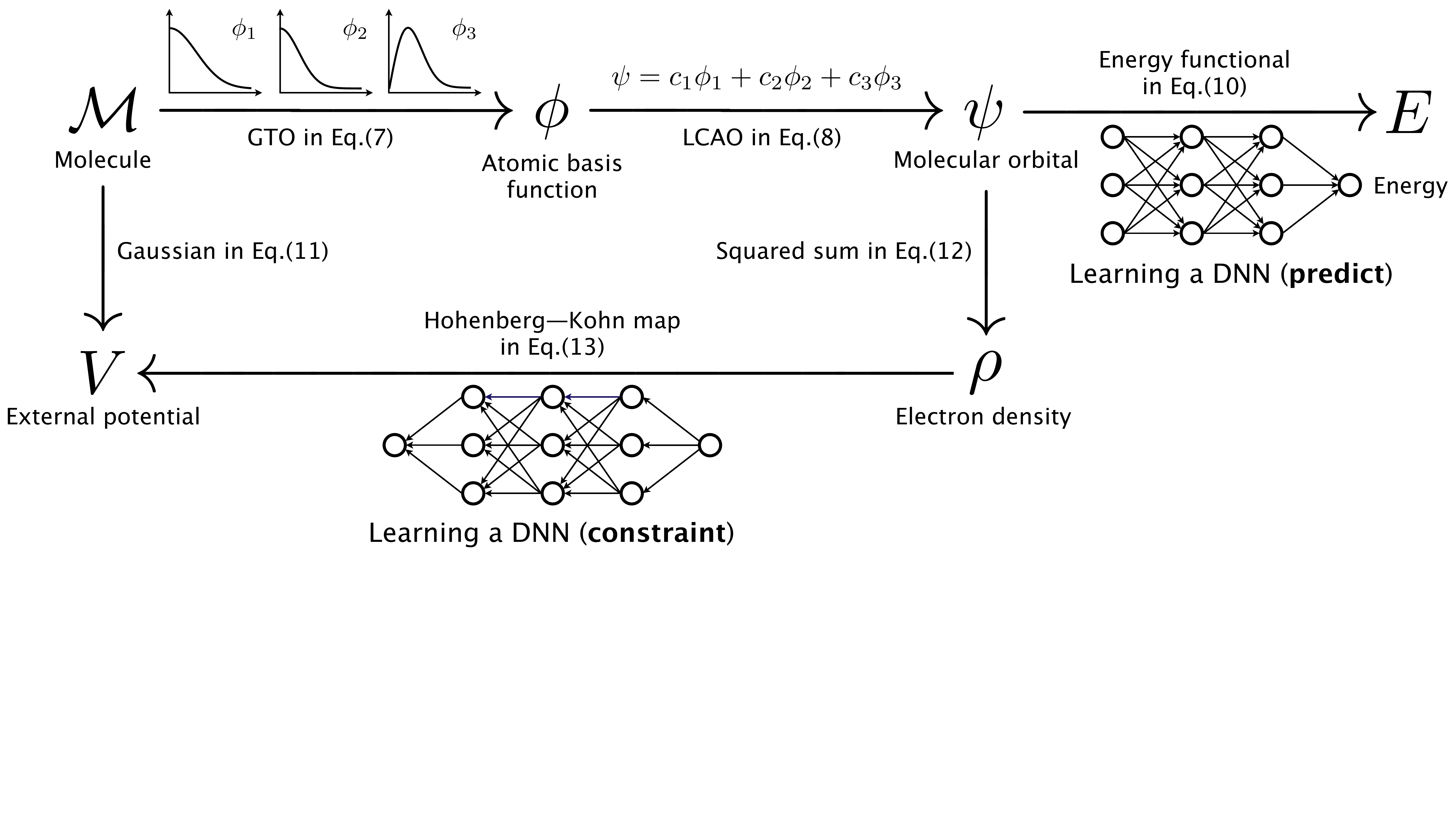}
\caption
{\label{overview}
Overview of the computational graph of our proposed quantum deep field (QDF)
framework from the input molecule $\mathcal M$ to the output energy $E$.
The energy functional predicts $E$ and the Hohenberg--Kohn (HK) map
imposes the potential constraint on the electron density $\rho$.
}
\end{center}
\end{figure}

From these observations, we describe the quantum deep field (QDF),
a machine learning (ML) model based on an underlying quantum physics,
in particular the density functional theory (DFT)~\cite{kohn1965self}.
The model is separated into linear and nonlinear components.
The former is the linear combination of atomic orbitals
(LCAO)~\cite{pauling1960nature,lennard1929electronic,mcquarrie1997physical},
which is implemented through matrix--vector multiplication;
the latter is the energy functional
that has nonlinearity inherent in quantum physics.
This study implements this nonlinear functional
using a vanilla feedforward DNN (Section~4.1).
Additionally, over the entire model, we impose a physical constraint
based on the Hohenberg--Kohn theorem~\cite{hohenberg1964inhomogeneous},
which has nonlinearity inherent in DFT and can therefore be
implemented using a vanilla feedforward DNN (Section~4.2).
The components and constraint can be represented as a computational graph that
learns the energy in a supervised fashion (Figure~\ref{overview}),
and all model parameters are trained by back-propagation
and stochastic gradient descent (SGD) algorithms (Section~4.3).
For atomization energy prediction with the QM9
dataset~\cite{ramakrishnan2014quantum}, our QDF model was competitive with
a state-of-the-art model called SchNet~\cite{schutt2018schnet}
but with a million fewer parameters (Section~5.1).

Furthermore, this study demonstrated an
``extrapolation''~\cite{smith2017ani,eickenberg2018solid}
with regard to predicting the energies of totally unknown molecules,
in which we trained a QDF model with small molecules, tested it with large molecules,
and achieved high extrapolation performance (Section~5.2).
In a standard ML evaluation,
the training and test sets have the same data distribution; in other words,
the molecular sizes and structures in both sets are the same or very similar.
Under this ``interpolation'' evaluation, if a highly nonlinear DNN model is trained,
it can easily fit to a physically meaningless but high-accuracy function
that maps the input molecules into the output energies; this is
because DNN can easily learn many non-linear properties inside the training data
distribution~\cite{xu2019can,xu2020neural} unrelated with physics.
However, the ML evaluation is mainly interested in the final output
(i.e., interpolation accuracy within a benchmark dataset),
and even if the energy prediction performs well,
the model parameters may not always reflect physics.
The extrapolation can evaluate ML models focusing on physical validity;
this will lead to the development of more reliable and practical ML applications.

\section{Background: molecular GCN and LCAO}

{\bf 2.1 Molecular GCN.}
A molecule is defined as $\mathcal M = \{ (a_1, R_1), (a_2, R_2),
\cdots, (a_M, R_M) \} = \{ (a_m, R_m) \}_{m=1}^{M}$,
where $a_m$ is the $m$th atom (e.g., \ce{H} and \ce{O}),
$R_m$ is the 3D coordinate of $a_m$,
and $M$ is the number of atoms in $\mathcal M$.
We consider a graph representation of $\mathcal M$, in which
the node is $a_m$ and the edge between $a_m$ and $a_n$
is defined by the atomic distance $D_{mn} = ||R_m - R_n||$.
In other words, the molecular graph $\mathcal G_{\mathcal M} = (V, D)$
is a fully connected graph, where $V$ is the set of atoms
and $D \in \mathbb R^{M \times M}$ is the corresponding distance matrix.

Given $\mathcal G_{\mathcal M}$, we initialize each atom with
a $d$-dimensional vector and denote the atom vector as
$\mathbf a_m$, where $d$ is a hyperparameter.
For example, $\mathbf a_m = [a_1, a_2, \cdots, a_d]$,
in which each element is a feature value
(e.g., the atomic charge, hybridization, and acceptor/donor)
used in~\cite{kearnes2016molecular,gilmer2017neural,chen2019graph},
or is randomly initialized and then learned
via back-propagation and SGD~\cite{schutt2017quantum,schutt2018schnet}.
We then consider the convolutional operation to update $\mathbf a_m$ iteratively
according to the graph structure $\mathcal G_{\mathcal M}$ as follows:
\begin{eqnarray}
\label{GCN}
\mathbf a_m^{(\ell+1)} = \sum_{n=1}^M w(D_{mn}) \mathbf h_n^{(\ell)},
\end{eqnarray}
where $\mathbf a_m^{(\ell+1)}$ is the $m$th atom vector in layer $\ell+1$
(where layer means the number of updates or iterations),
$\mathbf h_n^{(\ell)}$ is the hidden vector of $\mathbf a_n^{(\ell)}$
obtained by a neural network (e.g., $\mathbf h_n^{(\ell)} =
\text{ReLU} (\mathbf W^{(\ell)} \mathbf a_n^{(\ell)} + \mathbf b^{(\ell)})$),
and $w(D_{mn})$ is a function of the weight of $\mathbf h_n^{(\ell)}$
determined by the atomic distance.
Various weight functions (or edge features) have been used; for example,
the chemical bond type~\cite{kearnes2016molecular,chen2019graph},
distance bin~\cite{gilmer2017neural},
inverse: $w(D_{mn}) = 1 / D_{mn}$,
Gaussian kernel transformation: $w(D_{mn})
= \exp(-\gamma ||D_{mn} - \mu||^2)$~\cite{schutt2017quantum,schutt2018schnet},
and learnable vector: $w(D_{mn})
= \mathbf u_{mn} \in \mathbb R^d$~\cite{xie2018crystal}.
Note that, if we consider the molecular graph
represented by the adjacency matrix $A$ instead of the distance matrix $D$,
we have the binary weight $w(D_{mn}) = \{ 0, 1 \}$ that corresponds to the bond.

{\bf 2.2 LCAO.}
Herein, for the readers who are not familiar with quantum physics and chemistry,
we start from a slightly incorrect but understandable description of LCAO
(also called the linear superposition of wave functions)
introduced by Pauling and Lennard-Jones in
the 1920s~\cite{pauling1960nature,lennard1929electronic,mcquarrie1997physical}.
We then provide a correct description to facilitate comparison of
the equivalence and differences between molecular GCN and LCAO in Section~3.

\begin{wrapfigure}{r}{6cm}
\begin{center}
\includegraphics[width=6cm, bb=0 400 1347 900]{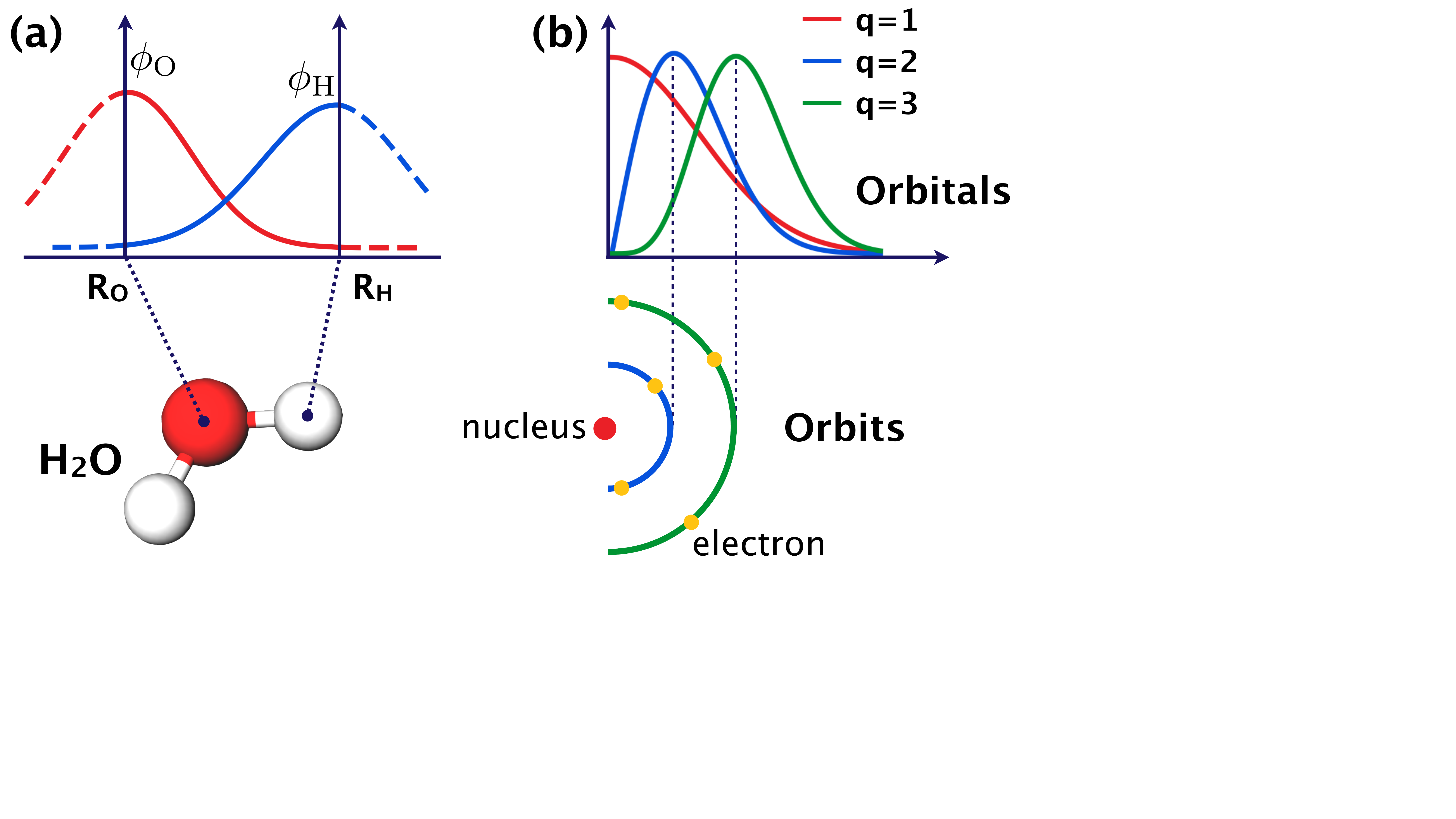}
\caption
{\label{orbitals/orbits}
(a)~Superposition between $\phi_\text{O}$ and $\phi_\text{H}$
corresponding to the chemical bond \ce{O-H} in \ce{H2O}.
(b)~``Orbitals'' in a quantum view and ``orbits'' in a classical view.
}
\end{center}
\end{wrapfigure}

We first consider that each atom $a_m$ has an inherent wave-like
spread in 3D space centered on $R_m$; this is known as the electron cloud.
As shown in Figure~\ref{orbitals/orbits}(a),
we can consider this wave as a probability distribution.
Based on the waves, we consider a value (i.e., the probability of electron)
on an arbitrary position $r$ in the ``field'' of $\mathcal M$,
not limited to the atomic position $R_m$, as follows:
\begin{eqnarray}
\psi(r) = \sum_{m=1}^{M} c_m \phi_m(r - R_m),
\end{eqnarray}
where $\psi(r)$ is the value on $r$, $\phi_m(r - R_m)$ is the value on $r$
derived from the wave whose origin is $R_m$,
and $c_m$ is the coefficient in this linear combination.
In quantum physics or chemistry, we refer to
$\phi$ as the atomic wave function or atomic orbital and
$\psi$ as the molecular wave function or molecular orbital.
For example, in a water molecule \ce{H2O},
the above linear combination can be expressed as follows:
\begin{eqnarray}
\label{H2O_1}
\psi(r) = c_\text{H} \phi_\text{H}(r - R_\text{H})
+ c_\text{H} \phi_\text{H}(r - R_{\text{H}'})
+ c_\text{O} \phi_\text{O}(r - R_\text{O}).
\end{eqnarray}
Note that because the two hydrogen atoms in \ce{H2O}
have the same characteristics, its atomic orbital and coefficient
are the same (i.e., $\phi_\text{H}$ and $c_\text{H}$),
but we distinguish their different positions
using $R_\text{H}$ and $R_{\text{H}'}$ in Eq.~(\ref{H2O_1}).

However, Eq.~(\ref{H2O_1}) is incorrect
because the oxygen atom \ce{O} has multiple electrons
and we need to consider multiple atomic orbitals for \ce{O}.
That is, we rewrite the term \ce{O} in Eq.~(\ref{H2O_1}) as follows:
\begin{eqnarray}
\label{H2O_2}
\psi(r) = c_\text{H} \phi_\text{H}(r - R_\text{H})
+ c_\text{H} \phi_\text{H}(r - R_{\text{H}'})
+ \sum_k c^{(k)}_\text{O} \phi^{(k)}_\text{O}(r - R_\text{O}),
\end{eqnarray}
where $k$ is the number of atomic orbitals for the oxygen.
Additionally, we have various choices for
representing each atomic orbital in Eq.~(\ref{H2O_2}).
Here, it is natural to represent an orbital
using a large number of known valid (e.g., Gaussian) basis functions.
That is, we further rewrite Eq.~(\ref{H2O_2}) using $\phi^{(\cdot)}$,
which is called the atomic basis function as follows:
\begin{eqnarray}
\label{H2O_3}
\psi(r) = \sum_i c^{(i)}_\text{H} \phi^{(i)}_\text{H}(r - R_\text{H})
+ \sum_i c^{(i)}_\text{H} \phi^{(i)}_\text{H}(r - R_{\text{H}'})
+ \sum_j \sum_k c^{(j,k)}_\text{O} \phi^{(j,k)}_\text{O}(r - R_\text{O}).
\end{eqnarray}
Thus, Eq.~(\ref{H2O_3}) has a nested structure, in which
the molecular orbital $\psi(r)$ has multiple atomic orbitals
$\phi_a(r - R_a)$ and each atomic orbital has multiple
atomic basis functions $\phi^{(\cdot)}_a(r - R_a)$.
Generally, we flatten this nested structure and describe it as follows:
\begin{eqnarray}
\label{LCAO}
\psi(r) = \sum_{n=1}^N c_n \phi_n(r - R_n)
\:\:\: \text{s.t.} \:\:\: \sum_{n=1}^N c_n^2 = 1,
\end{eqnarray}
where $N$ is the total number of basis functions
and the coefficients are normalized.
Eq.~(\ref{H2O_1}) has only three terms when $\mathcal M = \text{\ce{H2O}}$,
whereas, Eq.~(\ref{LCAO}) can have a large number of terms
(in principle, infinite) for better approximating $\psi$.
The set of basis functions (i.e., the basis set)
determines the level of computational accuracy
in theoretical calculations or quantum chemical simulations.
For example, with a standard basis set, such as
6-31G~\cite{mcquarrie1997physical} that is widely used in theoretical calculations,
\ce{H2O} has more than 30 basis functions (i.e., $N > 30$)
and \ce{C6H6} (benzene) has more than 100 basis functions (i.e., $N > 100$).

In addition to the number of basis functions,
the types of each basis function are important.
The theoretical calculations often use the Gaussian-type orbital (GTO) as follows:
\begin{eqnarray}
\label{GTO}
\phi_n(r - R_n) = \frac{1}{Z(q_n, \zeta_n)} D_n^{(q_n-1)} e^{-\zeta_n D_n^2},
\end{eqnarray}
where $D_n = ||r - R_n||$, $q_n = 1, 2, \cdots$ is the principle quantum number,
$\zeta_n$ is the control parameter of the Gaussian expansion
(called the orbital exponent), and $Z(q_n, \zeta_n)$ is the normalization term.
Note that this GTO is simplified in terms of the spherical harmonics.
In Eq.~(\ref{GTO}), $D_n^{(q_n-1)}$ allows the orbital to shift
the peak of the Gaussian expansion, which corresponds to
the classical ``orbit'' as shown in Figure~\ref{orbitals/orbits}(b).
This is because the wave function is referred to as the ``orbital.''

\begin{figure}[t]
\begin{center}
\includegraphics[width=14cm, bb=0 200 1920 900]{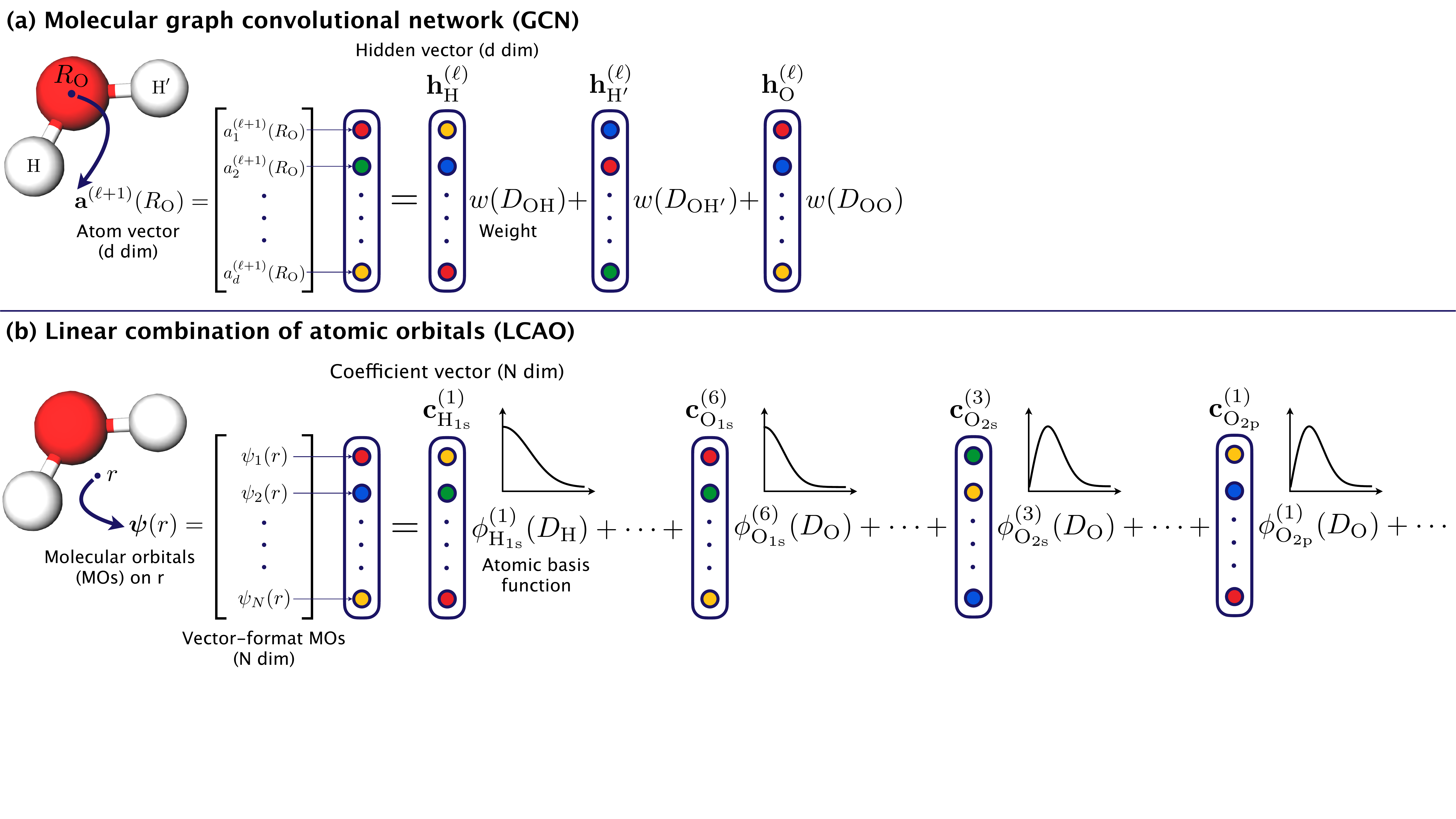}
\caption
{\label{GCNvsLCAO}
Both calculations involve the summation of the vector--scalar multiplications.
(a)~In the molecular GCN, the vectors are the atomic features
and the scalars are their weights.
(b)~In the LCAO, the vectors are the coefficients
and the scalars are the values of the atomic basis functions on $r$.
Roughly, with a standard 6-31G basis set,
we have six basis functions for the 1s orbital of the oxygen,
which we denote as $\phi_{\text{O}^{(1)}_\text{1s}},
\phi_{\text{O}^{(2)}_\text{1s}},
\cdots, \phi_{\text{O}^{(6)}_\text{1s}}$.
Additionally, we have four (i.e., $3+1=4$) basis functions for
the 1s orbital of the hydrogen and the 2s/2p orbitals of the oxygen,
which we denote as $\phi_{\text{H}^{(1)}_\text{1s}},
\cdots, \phi_{\text{H}^{(4)}_\text{1s}}$,
$\phi_{\text{O}^{(1)}_\text{2s}}, \cdots, \phi_{\text{O}^{(4)}_\text{2s}}$, and
$\phi_{\text{O}^{(1)}_\text{2p}}, \cdots, \phi_{\text{O}^{(4)}_\text{2p}}$.
Each $N$-dimensional coefficient vector is also characterized by a subscript
(e.g., $\mathbf c_{\text{H}^{(1)}_\text{1s}}$,
$\mathbf c_{\text{O}^{(6)}_\text{1s}}$,
$\mathbf c_{\text{O}^{(3)}_\text{2s}}$,
and $\mathbf c_{\text{O}^{(1)}_\text{2p}}$).
}
\end{center}
\end{figure}

We finally provide the correct description of LCAO.
Indeed, the LCAO has $N$ multiple molecular orbitals
and represents the linear combination using
the coefficient vector $\mathbf c_n \in \mathbb R^N$ as follows:
\begin{eqnarray}
\label{LCAOvector}
\boldsymbol \psi(r) = \sum_{n=1}^N \mathbf c_n \phi_n(r - R_n),
\end{eqnarray}
where $\boldsymbol \psi(r) \in \mathbb R^N$
is the vector-format molecular orbitals on $r$
and its $n$th element (i.e., $\psi_n(r)$) is the $n$th molecular orbital.
Note that an initial assumption in LCAO is that the number of molecular orbitals
is equal to the number of atomic orbitals (or basis functions).
In other words, the vector dimensionality $N$
and the number of basis functions $N$ are the same.
Thus, as we enhance the computational accuracy for approximating
$\boldsymbol \psi(r)$ by increasing the number of basis functions,
the dimensionality of $\boldsymbol \psi(r)$ also increases.

\section{Equivalence and difference between molecular GCN and LCAO}

Herein, in the molecular GCN, we rewrite the left side of Eq.~(\ref{GCN}) as
$\mathbf a_m^{(\ell+1)} = \mathbf a^{(\ell+1)}(R_m)$
because the atom vector is defined on the atomic position $R_m$.
Additionally, we transpose two terms
$w(D_{mn}) \mathbf h_n^{(\ell)}$ on the left side of Eq.~(\ref{GCN})
(i.e., the weight--vector) to $\mathbf h_n^{(\ell)} w(D_{mn})$ (i.e., the vector--weight).
In the practical calculation of LCAO, we create a grid field of $\mathcal M$
and position $r$ in continuous space is regarded as a discrete
point $r_i$ in the grid field (see Supplementary Material).
Additionally, we rewrite the right side of Eq.~(\ref{LCAOvector})
as $\phi_n(r_i - R_n) = \phi_n(D_{in})$ because the atomic basis function
is a function of the distance $D_{in} = ||r_i - R_n||$,
except for the other parameters $q_n$ and $\zeta_n$.

Therefore, Eq.~(\ref{GCN}) and Eq.~(\ref{LCAOvector}) can be represented as follows:
\begin{eqnarray}
\label{GCN--LCAO}
\mathbf a^{(\ell+1)}(R_m) = \sum_{n=1}^M \mathbf h^{(\ell)}_n w(D_{mn})
\:\:\:\:\:\: \text{and} \:\:\:\:\:\:
\boldsymbol \psi(r_i) = \sum_{n=1}^N \mathbf c_n \phi_n(D_{in}).
\end{eqnarray}
Thus, the two equations are easier to compare (Figure~\ref{GCNvsLCAO}).
In the following subsections, we find and discuss
the equivalence and difference between the molecular GCN and LCAO.

{\bf 3.1 Atom vector $\mathbf a^{(\ell+1)}(R_m) \in \mathbb R^d$
and molecular orbital $\boldsymbol \psi(r_i) \in \mathbb R^N$.}
On the left side of both equations in Eq.~(\ref{GCN--LCAO}),
the GCN has the $d$-dimensional vector on the atomic position $R$,
and the LCAO has the $N$-dimensional vector on the field position $r_i$.
We believe that this difference, in terms of the positions of
the multidimensional variables, is not a serious problem in this case.
We can assume $R$ as a representative point that the GCN considers.
Actually, this is reasonable for modeling the molecule and its energy
because the molecular energy can often be calculated as
the summation of atomic energy contributions~\cite{behler2007generalized}.
For example, in physical chemistry the neural network
potential~\cite{behler2007generalized,behler2011atom,smith2017ani}
based on the embedded atom method~\cite{daw1984embedded} has been proposed;
this method considers the LCAO, molecular orbital, and electron density
on the atomic position $R$, and not on the field position $r_i$~\cite{zhang2019embedded}.
Furthermore, $d$ is a hyperparameter in the GCN and
$N$ is the number of basis functions in the LCAO, which can be varied.
When $d = N$, both the GCN and LCAO have information
with the same expressive power on a position in 3D space.

{\bf 3.2 Hidden vector $\mathbf h^{(\ell)}_n \in \mathbb R^d$
and coefficient vector $\mathbf c_n \in \mathbb R^N$.}
On the right side of both equations in Eq.~(\ref{GCN--LCAO}),
the GCN has the hidden (i.e., nonlinear transformed atom) vector
$\mathbf h^{(\ell)}_n = \text{ReLU}
(\mathbf W^{(\ell)} \mathbf a^{(\ell)}_n + \mathbf b^{(\ell)})$,
which often contains various atomic features
(e.g., the charge, hybridization, and acceptor/donor) in its
elements~\cite{kearnes2016molecular,gilmer2017neural,chen2019graph}.
On the other hand, the LCAO has the coefficient vector $\mathbf c_n$
and we can find that $\mathbf h^{(\ell)}_n$ corresponds to $\mathbf c_n$.
However, $\mathbf c_n$ does not contain the atomic features and only has
the normalization condition $\sum_{n=1}^N c_n^2 = 1$ in Eq.~(\ref{LCAO}).
Although $\mathbf h^{(\ell)}_n$ and $\mathbf c_n$ have such different
characteristics, if both vectors are optimized for predicting/minimizing
the energy of a molecule, their role is the same.
Furthermore, we emphasize that the number of parameters and model complexity
of the GCN are significantly greater than those of the LCAO,
which is derived from $\mathbf W^{(\ell)}$, a nonlinear activation $\text{ReLU}$,
and a $\ell$ times iterative procedure in the GCN.

\begin{figure}[t]
\begin{center}
\includegraphics[width=14cm, bb=0 450 1920 1000]{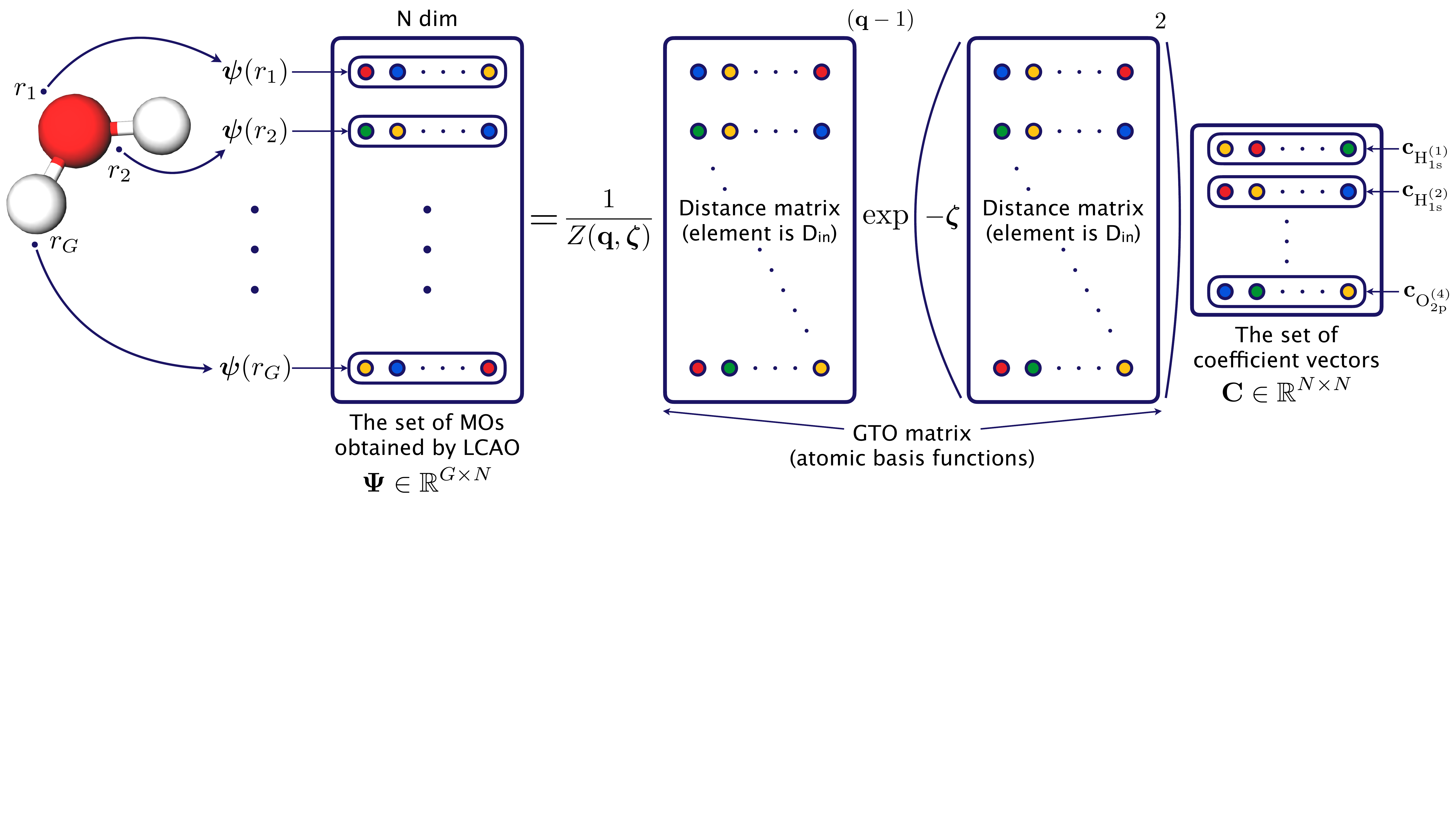}
\caption
{\label{batchLCAO}
Intuitive explanation of the linear combination of atomic orbitals
(LCAO) using batch-processing based matrix--vector multiplication (e.g., in PyTorch).
The set of molecular orbitals on all points in the gird field,
$\boldsymbol \Psi = \{ \boldsymbol \psi(r_i) \}_{i=1}^G$, is efficiently obtained
(i.e., processed as a batch on a GPU) by matrix-vector multiplication
between the Gaussian-type orbital (GTO) matrix
(i.e., each element is the atomic basis function $\phi(D_{in})$)
and the set of coefficient vectors $\mathbf C$.
}
\end{center}
\end{figure}

{\bf 3.3 Weight $w(D_{mn}) \in \mathbb R$
and basis function $\phi_n(D_{in}) \in \mathbb R$.}
In Eq.~(\ref{GCN--LCAO}), we have the weight $w(D_{mn})$
on the hidden vector in the GCN and find that $w(D_{mn})$
corresponds to the basis function $\phi_n(D_{in})$ in the LCAO.
We emphasize that the $d$-dimensional hidden vectors are
summed and weighted by the distances in the GCN;
in contrast, the basis functions by the distances are combined linearly
with the $N$-dimensional coefficient vectors in the LCAO.
For example, in a GCN model, if the weight function is a Gaussian kernel
transformation (e.g., $w(D_{mn}) = \exp((D_{mn} - \mu)^2 / \sigma^2)$),
it has similar characteristics to the GTO; however,
this does not include the term $D_{in}^{(q_n-1)}$ in Eq.~(\ref{GTO}).
Therefore, this does not satisfy the original symmetry
and cannot represent the different peaks in Figure~\ref{orbitals/orbits}(b).
In particular, when a molecule is represented as a discrete graph
using an adjacency matrix (i.e., $w(D_{mn}) = \{ 0, 1 \}$),
the atomic basis function (or orbital/wave function) is $0$ or $1$.

{\bf 3.4 Number of atoms $M$ and number of basis functions $N$.}
Let us focus on the number of terms in Eq.~(\ref{GCN--LCAO}).
We can find that the GCN has $M$ number of terms (atoms),
where, for example, $M = 3$ when $\mathcal M = \text{\ce{H2O}}$.
In contrast, the LCAO has $N$ number of terms (basis functions),
where, for example, $N > 30$ when $\mathcal M = \text{\ce{H2O}}$
using a standard basis set (e.g., 6-31G, as mentioned in Section~2.2).
The difference is more than 10 times; that is, the GCN has a very poor basis set.
Recent studies have pointed out that GCNs cannot improve their performance
by increasing the number of convolutional layers~\cite{kipf2016semi,xu2019powerful}.
While we are not familiar with GCN applications for other kinds of graph data,
such as social, biological, and financial networks,
for the molecular GCN and its variants, the above problem is trivial:
the number of basis functions is insufficient
and the type of each basis function is incorrect.
The overparameterized, deeply hierarchical, and highly nonlinear GCN model
builds on such poor and incorrect basis sets.
Note that while the molecular GCN uses the $M \times M$ distance matrix,
the LCAO uses the $G \times N$ GTO matrix, where $N$ is larger than $M$
and $G$ is much larger than $M$ (Figure~\ref{batchLCAO}).
This is inevitable in the LCAO but a computational drawback
(in particular, for the memory) compared with the GCN.

{\bf 3.5 Dimensionality of the atom vector $d$
and number of basis functions $N$.}
The dimensionality $d$ of the atom vector in the GCN corresponds to
the number of atomic features;
the dimensionality $N$ of the coefficient vector in the LCAO corresponds to
the number of basis functions, which determines the level of computational accuracy
in theoretical calculations or quantum chemical simulations.
Considering this along with Section 3.4, although it seems to enhance
the expressive power of the GCN by increasing $d$,
this is not physically meaningful owing to the fixed $M$,
where $M$ is both the molecular size and the number of basis functions.
For example, even if we use the dimensionality $d = 1000$
when $\mathcal M = \text{\ce{H2O}}$,
the number of basis functions is still only three in the GCN.
In contrast, in LCAO, $N$ is not only the dimensionality
of coefficient vector but also the number of basis functions.
Therefore, increasing $N$ is physically meaningful for approximating
$\boldsymbol \psi(r_i)$ and improving the computational accuracy.
Furthermore, $N$ is usually determined by a basis set;
therefore, we can assume that $N$ is an automatically determined value
and not a hyperparameter, like in ML models.

A molecular GCN is regarded as
a neural message passing algorithm~\cite{gilmer2017neural}, in which
the atom is an object (i.e., a node) that has a message
(i.e., a feature vector), and the atom messages are passed through
the molecular graph structure using the adjacency or distance matrix
in a nonlinear deep fashion; however, this is incorrect.
More precisely, the atomic orbital or wave function is represented by
the basis functions, and their linear combination or superposition is calculated using
the multidimensional coefficients; this is the LCAO (see again Figure~\ref{GCNvsLCAO}).
Table~1 summarizes the characteristics of
the molecular GCN and LCAO discussed in this section.

\section{Learning energy functional and imposing physical constraint}

{\bf 4.1 Energy functional.}
Thus far, we have described the linearity
in obtaining molecular orbitals from atomic orbitals.
However, the relationship between molecular orbitals and energy
has nonlinearity inherent in quantum physics.
Here, we use a neural network to estimate this nonlinear relationship,
which is called an energy functional
(i.e., function $\mathcal F$ of function $f(x)$, $\mathcal F[f(x)]$)
because the energy $E$ is a function of $\psi$ and $\psi$ is a function of $r$.

As shown in Figure~\ref{batchLCAO},
we have the set of $N$-dimensional vector-format molecular orbitals:
$\boldsymbol \Psi = \{ \boldsymbol \psi(r_i) \}_{i=1}^G$.
We consider $\mathcal{F}_\text{DNN}$,
which is a DNN-based energy functional as follows:
\begin{eqnarray}
\label{FDNN}
E'_{\mathcal M} = \mathcal{F}_\text{DNN} [\boldsymbol \Psi],
\end{eqnarray}
where $E'_{\mathcal M}$ is the predicted energy of $\mathcal M$.
This study uses a vanilla feedforward DNN for the implementation of
$\mathcal{F}_\text{DNN}$ (see Supplementary Material).
We finally minimize the loss function:
$\mathcal L_E = || E_{\mathcal M} - E'_{\mathcal M}||^2$,
where $E_{\mathcal M}$ is the actual energy provided by the training dataset.

{\bf 4.2 Physical constraint based on the Hohenberg--Kohn theorem.}
Unfortunately, only minimizing $\mathcal L_E$
does not result in a physically meaningful learning model.
Since the DNN-based energy functional has strong nonlinearity,
$\mathcal{F}_\text{DNN}$ may output the actual energy
even if its input molecular orbitals $\boldsymbol \Psi$ are incorrect.
To address this, we impose a physical constraint in learning the model
based on the Hohenberg--Kohn theorem~\cite{hohenberg1964inhomogeneous},
which ensures that ``the external potential $V(r)$
is a unique functional of the electron density $\rho(r)$.''
$V(r)$ can be determined by the atomic charges $Z_m$
(e.g., $Z_{\ce{H}}=1$ and $Z_{\ce{O}}=8$) and their positions $R_m$.
Additionally, $\rho(r)$ can be obtained
by\footnote{This density is based on the one electron approximation.
Precisely, $\boldsymbol \psi(r)$ is not the true molecular orbital
and is called the Kohn--Sham orbital~\cite{kohn1965self},
which is a molecular orbital in a fictitious system of non-interacting electrons
that provides the same density as the system of interacting electrons.}
$\rho(r) = \sum_{n=1}^N |\psi_n(r)|^2$.
Furthermore, the aforementioned statement indicates that
a relationship between $V(r)$ and $\rho(r)$
has a nonlinearity but one-to-one correspondence inherent in DFT.
Here, we also use a neural network to estimate this nonlinear relationship;
this is called a Hohenberg--Kohn (HK) map~\cite{brockherde2017bypassing,moreno2020deep}.
The HK map works as the constraint on $\boldsymbol \psi(r)$,
which is the input of $\mathcal{F}_\text{DNN}$, and results in a physically meaningful
learning model as a whole (see again Figure~\ref{overview}).

Formally, we first consider a Gaussian-based external
potential~\cite{bartok2010gaussian,brockherde2017bypassing} on $r_i$ as follows:
\begin{eqnarray}
\label{V}
V_{\mathcal M}(r_i) = - \sum_{m=1}^M Z_m e^{-||r_i - R_m||^2}.
\end{eqnarray}
Thus, this is the Gaussian expansions of atomic charges in 3D space.
Note that the model assumes $V_{\mathcal M}(r_i)$
to be the correct external potential of $\mathcal M$; that is,
$V_{\mathcal M}(r_i)$ is used as a target for minimizing loss in the model.
Additionally, we have the electron density on $r_i$ as
\begin{eqnarray}
\label{density}
\rho(r_i) = \sum_{n=1}^N | \psi_n(r_i) |^2
\end{eqnarray}
and consider $\mathcal{HK}_\text{DNN}$, which is a DNN-based HK map as follows:
\begin{eqnarray}
\label{HKDNN}
V'_{\mathcal M}(r_i) = \mathcal{HK}_\text{DNN}(\rho(r_i)),
\end{eqnarray}
where $V'_{\mathcal M}(r_i)$ is the predicted external potential of $\mathcal M$.
This study uses a vanilla feedforward DNN for the implementation of
$\mathcal{HK}_\text{DNN}$ (see Supplementary Material).
We finally minimize the loss function:
$\mathcal L_V = ||V_{\mathcal M}(r_i) - V'_{\mathcal M}(r_i)||^2$.

{\bf 4.3 Learning.}
As a total learning model, we minimize $\mathcal L_E$ for predicting $E$
and $\mathcal L_V$ for imposing the potential constraint on $\rho$ ``alternately.''
We believe that this is similar to the learning strategy of
generative adversarial networks (GANs)~\cite{goodfellow2014generative}.
The entire model optimizes all parameters in
the LCAO, $\mathcal{F}_\text{DNN}$, and $\mathcal{HK}_\text{DNN}$
using only the back-propagation and SGD algorithms.
This is our proposed QDF framework
and we describe its optimization details in Supplementary Material.

It is important to note that QDF must consider the following physical condition:
\begin{eqnarray}
N_\text{elec} = \int \rho(r) dr
= \int \sum_{n=1}^N |\psi_n(r)|^2 dr
\approx \sum_{i=1}^G \sum_{n=1}^N |\psi_n(r_i)|^2,
\end{eqnarray}
where $N_\text{elec}$ is the total electrons in $\mathcal M$
(e.g., when $\mathcal M = \text{\ce{H2O}}$, $N_\text{elec} = 10$).
In other words, we must ``keep'' the total electrons in learning/updating
the molecular orbitals with the iterative SGD algorithm.
We implement this by transforming $\boldsymbol \psi_n$ in the SGD as follows:
\begin{eqnarray}
\boldsymbol \psi_n \leftarrow \sqrt{\frac{N_{\text{elec}}}{N}}
\frac{\boldsymbol \psi_n}{|\boldsymbol \psi_n|}.
\end{eqnarray}
Note that $\boldsymbol \psi_n \in \mathbb R^G$ is not the $n$th row vector
but the $n$th column vector of $\boldsymbol \Psi$ described in Figure~\ref{batchLCAO}.
The model requires other physical normalizations in learning
(see Supplementary Material).

\begin{table}[t]
    \begin{tabular}{cc}

        \begin{minipage}[c]{0.5\hsize}
            \centering
            \begin{tabular}{|l|r|r|}
                \hline
                & GCN & LCAO \\
                \hline \hline
                \# of terms & $M (\text{fixed})$ & $N \sim \infty$ \\
                \hline
                weight/basis & $w(D)$ & $D^{(q-1)} e^{-\zeta D^2}$ \\
                \hline
                vector dim & $d$ & $N \sim \infty$ \\
                \hline
                parameters & $\mathbf W^{(\ell)}$, $\mathbf b^{(\ell)}$, $\mathbf a$ & $\mathbf c$ \\
                \hline
                nonlinearity & e.g., ReLU & nothing \\
                \hline
            \end{tabular}
            \label{characteristics}
            \caption{Characteristics of the molecular graph convolutional network (GCN)
            and the linear combination of atomic orbitals (LCAO).}
        \end{minipage} &

        \begin{minipage}[c]{0.5\hsize}
            \centering
            \begin{tabular}{lrr}
                \hline \hline
                Model & Size & MAE \\
                \hline
                GCN & 483,631 & 1.58 \\
                DTNN~\cite{schutt2017quantum} & --- & 1.51 \\
                SchNet~\cite{schutt2018schnet} & 1,676,133 & 1.23 \\
                QDF & 495,262 & 1.21 \\
                \hline
                Chemical accuracy & & 1.00 \\
                \hline \hline
            \end{tabular}
            \label{modelsize}
            \caption{Model size and mean absolute error (MAE; kcal/mol)
                     on the QM9under14atoms dataset.}
        \end{minipage}
    \end{tabular}
\end{table}

\section{Evaluation: prediction and extrapolation}

{\bf 5.1 Energy prediction.}
We first describe the prediction performance
for the atomization energy at 0 K of the QM9under14atoms dataset,
which is a subset of the QM9 dataset~\cite{ramakrishnan2014quantum}
(the dataset and training details are given in Supplementary Material).
Table~2 shows the model sizes and final prediction errors
(mean absolute error (MAE), lower is better) of the baseline GCN, proposed QDF,
and other methods~\cite{schutt2017quantum,schutt2018schnet} as references.
SchNet, which is a variant of the deep tensor neural network (DTNN)
proposed earlier, is a standard state-of-the-art deep learning model.
We argue that the GCN is not a weak baseline because
it achieves a reasonable performance that is similar to that of the DTNN.
Additionally, we believe that the QDF outperforms
or competes with SchNet in terms of the prediction error.
However, in terms of the model size,
SchNet has more than 1.5 million learning parameters.
In contrast, QDF, which has less than
half a million parameters, is much more compact.
We note that these results can vary with the careful tuning of
its hyperparameters and SchNet may outperform QDF;
however, our main aim in this study is not to build a competitive model
with regard to ``interpolation'' performance within a single benchmark dataset.

\begin{figure*}[t]
\begin{center}
\includegraphics[width=14cm, bb=0 675 1920 1000]{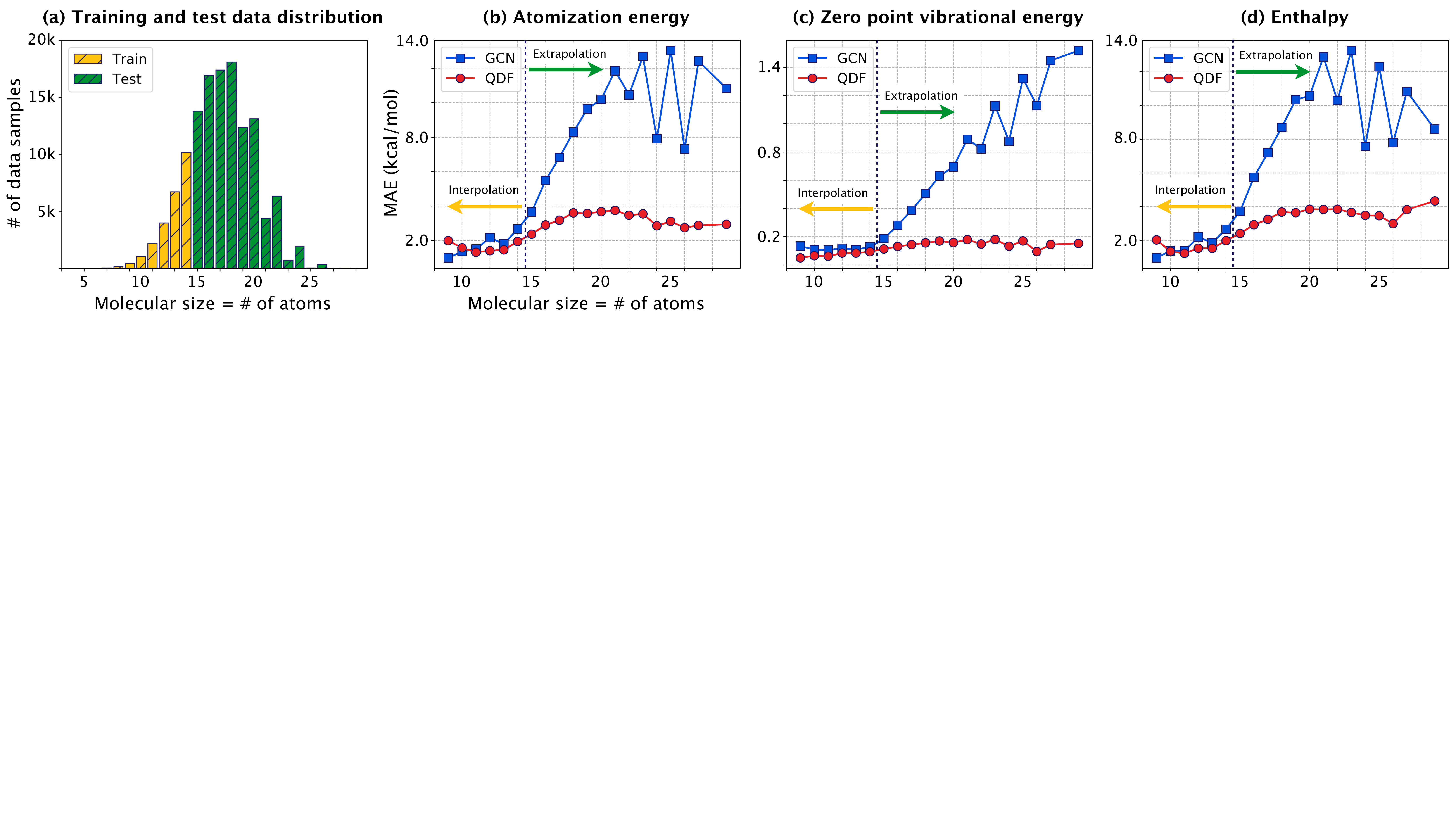}
\caption
{\label{extrapolation}
(a) Data distribution of training (interpolation) and test (extrapolation) samples.
The number of test samples of the QM9over15atoms dataset is 115k molecules,
which is 10 times more than that of training samples of the QM9under14atoms dataset.
(b), (c), and (d) are the mean absolute errors (MAEs) of the atomization energy,
zero point vibrational energy, and enthalpy of this large-scale extrapolation, respectively.
}
\end{center}
\end{figure*}

{\bf 5.2 Energy extrapolation.}
Using a more practical evaluation setting, we demonstrate that
the QDF can capture the physically meaningful energy functional.
In physics, we can assume that if an ML model could capture fundamental quantum
characteristics (i.e., the orbital/wave function $\psi$ and electron density $\rho$),
the model would be able to conduct a prediction for totally unknown molecules.
In other words, the QDF would be able to perform an ``extrapolation''
in predicting the energy of totally different sized and structured molecules.
To substantiate this, we trained a model with small molecules
consisting of fewer than 14 atoms (QM9under14atoms) and tested it with large molecules
consisting of more than 15 atoms (QM9over15atoms) in the QM9 dataset
(Figure~\ref{extrapolation}(a)).
Additionally, we used three kinds of energy properties provided by the QM9 dataset:
the atomization energy at 0 K, zero point vibrational energy, and enthalpy at 298.15 K.
As shown in Figure~\ref{extrapolation}(b), (c), and (d),
in each energy prediction, the GCN achieved accuracy comparable (or superior)
to that of the QDF in interpolation; however, the QDF could maintain this accuracy
even when the molecular size increased, whereas the GCN could not.
We believe that these low MAEs in predicting unknown large molecules
are evidence that the QDF can capture the physically meaningful energy functional
and the fundamental quantum characteristics of molecules.
Thus, we should move away from the competition of interpolation accuracy
and evaluate ML models using an extrapolation setting.

However, we also believe that the extrapolation performance for
the atomization energy and enthalpy have room to improve.
The current QDF uses the simplified GTO ignoring
the spherical harmonics~\cite{eickenberg2018solid},
simplified Gaussian external potential~\cite{brockherde2017bypassing}
that cannot reproduce the potential and density close to the nucleus,
and two vanilla feedforward DNNs.
To improve extrapolation performance, improvements to these factors
(e.g., using Slater-type orbitals (STOs), not GTOs) will be required.

In terms of computational cost, although the LCAO requires the $G \times N$ matrix
described in Figure~\ref{batchLCAO} and Section~3.4,
training a QDF model using 10k molecules of the QM9under14atoms dataset
can be done within 6 hours on a standard single (e.g., GTX 1080Ti) GPU.
Once the model is trained, the prediction for 100k molecules
of the QM9over15atoms dataset can be done within a few minutes.

\section{Discussion, conclusion, and future directions}
Our current implementation of the QDF model has some limitations.
In this implementation, orbital exponents $\{ \zeta_n \}_{n=1}^N$
and coefficient vectors $\{ \mathbf c_n \}_{n=1}^N$ in the LCAO
are the global common learning parameters for all molecules
(i.e., are not specific to each molecular structure).
These parameters actually differ for each real molecule
in theoretical calculations or quantum chemical simulations,
and this is considered in standard GCNs via
iterative nonlinear convolutional operations.
However, we believe that our global parameters and the linearity prevent
the model from being too flexible for each molecule in the training dataset;
this is a general and practical trade-off problem in optimizing a model.
Considering this trade-off, the current implementation of QDF is
reasonable in terms of reducing the model parameters and complexity,
resulting in robust extrapolation.
Additionally, $N$ is also different for each real molecule; however,
ML models need to set a global $N$ common for all molecules in a dataset.
Indeed, the QM9 dataset includes only small organic molecules
comprising H, C, N, O, and F atoms, and the maximum molecular size $M$ is 29.
Considering these and a standard basis set called 6-31G,
we set $N = 200$ as a valid value in theoretical calculations
(for other hyperparameters, see Supplementary Material).
Furthermore, this study considers an extrapolation evaluation setting
in terms of the number of atoms; however, some studies have demonstrated
extrapolation in terms of heavy atoms~\cite{smith2017ani,eickenberg2018solid}.
It is important to design various extrapolations
using various much larger datasets (e.g., the ANI-1~\cite{smith2017ani}
and Alchemy~\cite{chen2019alchemy}), which will require more efficient
implementation and learning of QDF with dozens of GPUs, in the future.

This study described the QDF framework, which is different from
other graph-based deep learning approaches for molecules; in other words,
QDF is not an extension of existing molecular GCN and SchNet models.
Our aim was to design a simple ML model consistently based on physics
without incorporating complicated DNN techniques/architectures.
Recently, some ML approaches have been proposed~\cite{grisafi2018transferable,
eickenberg2018solid,zhang2019embedded,chandrasekaran2019solving,
schutt2019unifying} as supervised learning models for
the orbital/wave function $\psi$ and electron density $\rho$.
In contrast, our QDF can be viewed as an unsupervised model to
reproduce $\rho$, originated from $\psi$,
using only a large-scale dataset of energy properties;
other training strategies, not the current GAN-like one (Section~4.3)
that may be unstable, will be considered in the future.

More generally, QDF can be viewed as one of the approaches
such as the physics-informed, Hamiltonian, Fermionic, and Pauli
neural networks~\cite{raissi2019physics,raissi2018hidden,raissi2018deep,
pun2019physically,greydanus2019hamiltonian,pfau2020ab,hermann2020deep};
these solve the physical problems and equations using physically meaningful modeling.
QDF is designed as a self-consistent learning machine for DFT
or to solve the Kohn--Sham equations~\cite{kohn1965self} with minimal (three)
components: LCAO, $\mathcal F_\text{DNN}$, and $\mathcal{HK}_\text{DNN}$.
We believe that integrating a supervised model
with a dataset of the electron density~\cite{sinitskiy2018deep}
(i.e., $\rho$ in Figure.~\ref{overview} is given as a target)
and an unsupervised but physically informed and meaningful model with a dataset
of the atomization energy, HOMO--LUMO gap, and other properties~\cite{chen2019alchemy}
will yield an interesting hybrid ML model.

In the future, QDF will allow extensions for not only molecules
but also for crystals~\cite{jain2013commentary,xie2018crystal,gong2019predicting}
and other practical applications in materials informatics.
In particular, the viability of extrapolation will lead to
the development of applications with transfer learning
for polymers~\cite{huan2016polymer}, catalysts~\cite{kitchin2018machine},
photovoltaic cells~\cite{lopez2016harvard,kim2017hybrid},
and fast, large-scale screening for discovering effective materials.
Extensions of the current QDF model can be created by forking our source code on GitHub.

\section*{Broader Impact}
This study will provide benefit for ML researchers who are interested in
quantum physics/chemistry and applications for materials science/informatics.

\begin{ack}
This study was supported by
the Grant-in-Aid for Early-Career Scientists (Grant No.~20K19876) from the JSPS
and the Grant-in-Aid for Scientific Research (Grant No.~19H05787 and 19H00818) from the MEXT.
\end{ack}

\bibliographystyle{unsrt}
\bibliography{references}

\newpage

\section*{Supplementary}

{\bf Dataset.}
The QM9 dataset~\cite{ramakrishnan2014quantum} contains
approximately 130,000 molecules made up of H, C, N, O, and F atoms
along with 13 quantum chemical properties
(e.g., atomization energy, HOMO, and LUMO) for each molecule.
These molecular properties were calculated using a hybrid quantum simulation
(Gaussian 09) at the B3LYP/6-31G(2df,p) level of theory.
In this study, we created a subset of the QM9 dataset
with a limited number of atoms, $M \leqq 14$, per molecule,
which we refer to as the ``QM9under14atoms'' dataset in the main text.
As the learning/predicting targets, we selected three kinds of energy properties:
atomization energy at 0 K, zero point vibrational energy, and enthalpy at 298.15 K.
The number of data samples in the QM9under14atoms dataset is approximately
15,000 molecules and we randomly shuffled and split this dataset into
training, development (or validation), and test sets with a ratio of 8:1:1,
in which the development set was used to tune the model and optimization hyperparameters.
For large-scale extrapolation evaluation, we tested the trained QDF model
using the ``QM9over15atoms'' dataset (i.e., $M \geqq 15$),
in which the number of data samples was approximately 115,000 molecules.

{\bf Molecular field definition.}
Given a molecule $\mathcal M = \{ (a_m, R_m) \}_{m=1}^M$,
we consider spheres with a radius of $s$~\AA,
where each sphere covers each atom centered on $R_m$, and then
divide the sphere into grids (or meshes) in intervals of $g$~\AA.
These $s$ and $g$ are the hyperparameters and
this process yields many grid points in $\mathcal M$.
The grid-based field of $\mathcal M$ is denoted by
$\{ r_1, r_2, \cdots, r_G \} = \{ r_i \}_{i=1}^G$,
where $r_i$ is the 3D coordinate of the $i$th point
and $G$ is the total number of points.

\begin{figure}[t]
\begin{center}
\includegraphics[width=14cm, bb=0 0 1920 1080]{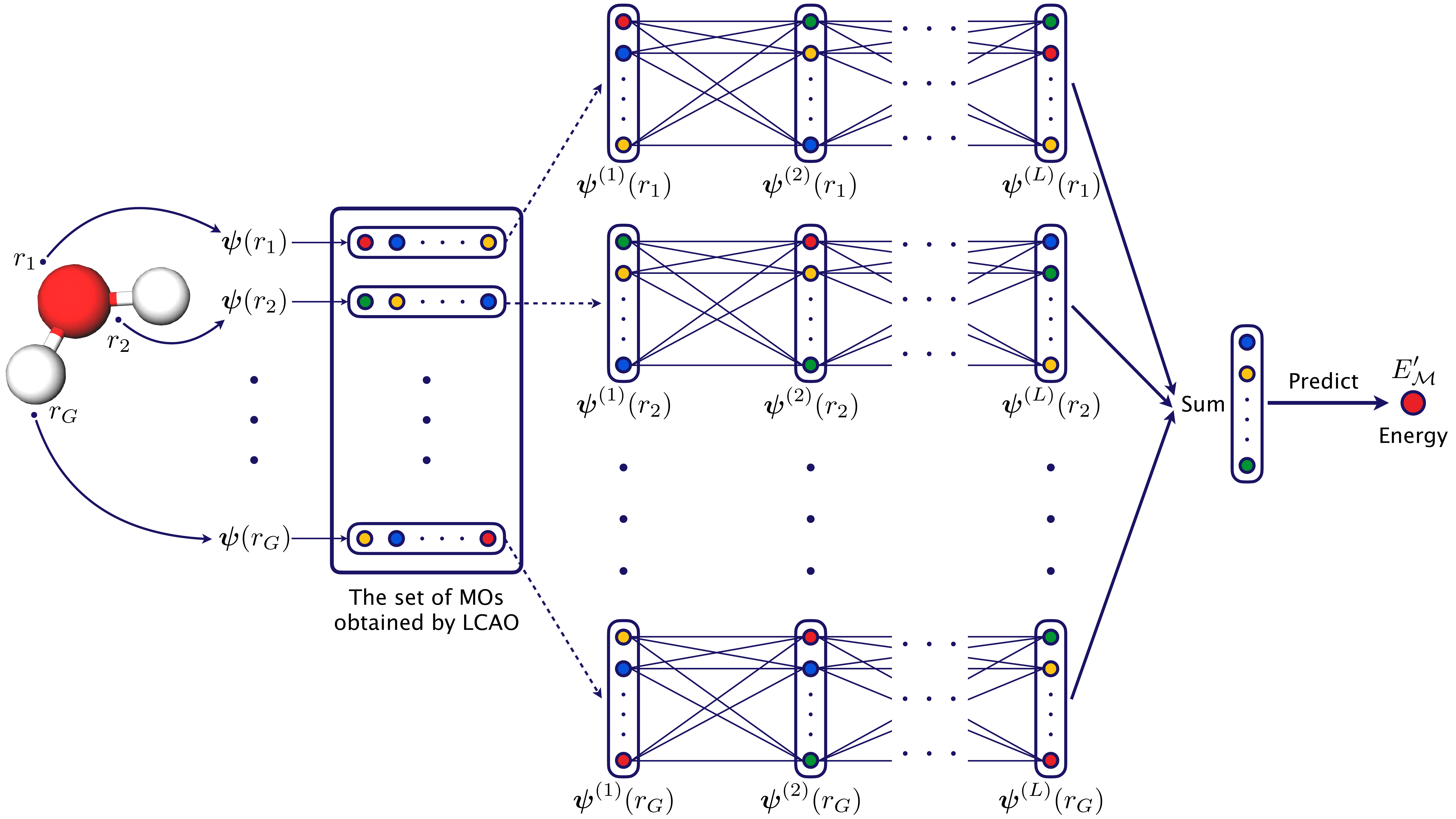}
\caption
{\label{fig_FDNN}
Architecture of our DNN-based energy functional $\mathcal{F}_{\text{DNN}}$.
}
\end{center}
\end{figure}

{\bf Architectures of $\mathcal{F}_\text{DNN}$ and $\mathcal{HK}_\text{DNN}$.}
To implement the DNN-based energy functional $\mathcal{F}_{\text{DNN}}$
in the main text, we consider the following vanilla feedforward architecture:
\begin{eqnarray}
\boldsymbol \psi^{(\ell+1)}(r_i) = \text{ReLU}
(\mathbf W_E^{(\ell)} \boldsymbol \psi^{(\ell)}(r_i) + \mathbf b_E^{(\ell)}),
\end{eqnarray}
where $\ell = 1, 2, \cdots, L$ is the number of hidden layers
($\boldsymbol \psi^{(1)}(r_i) = \boldsymbol \psi(r_i)$ and $L$ is the final layer),
ReLU is the nonlinear activation function,
$\mathbf W_E^{(\ell)} \in \mathbb R^{N \times N}$ is the weight matrix in layer $\ell$,
and $\mathbf b_E^{(\ell)} \in \mathbb R^N$ is the bias vector in layer $\ell$.
We then sum over $\{ \boldsymbol \psi^{(L)}(r_i) \}_{i=1}^{G}$
in the final layer $L$ and output an energy
(i.e., atomization energy, zero point vibrational energy, or enthalpy)
with the following vanilla linear regressor:
\begin{eqnarray}
E'_{\mathcal M} = \mathbf w_E^{\top}
\Big( \sum_{i=1}^{G} \boldsymbol \psi^{(L)}(r_i) \Big) + b_E,
\end{eqnarray}
where $\mathbf w_E \in \mathbb R^N $ is the weight vector
and $b_E \in \mathbb R$ is the bias scalar.
Figure~\ref{fig_FDNN} illustrates the architecture of this DNN-based energy functional.

\begin{figure}[t]
\begin{center}
\includegraphics[width=14cm, bb=0 0 1920 1080]{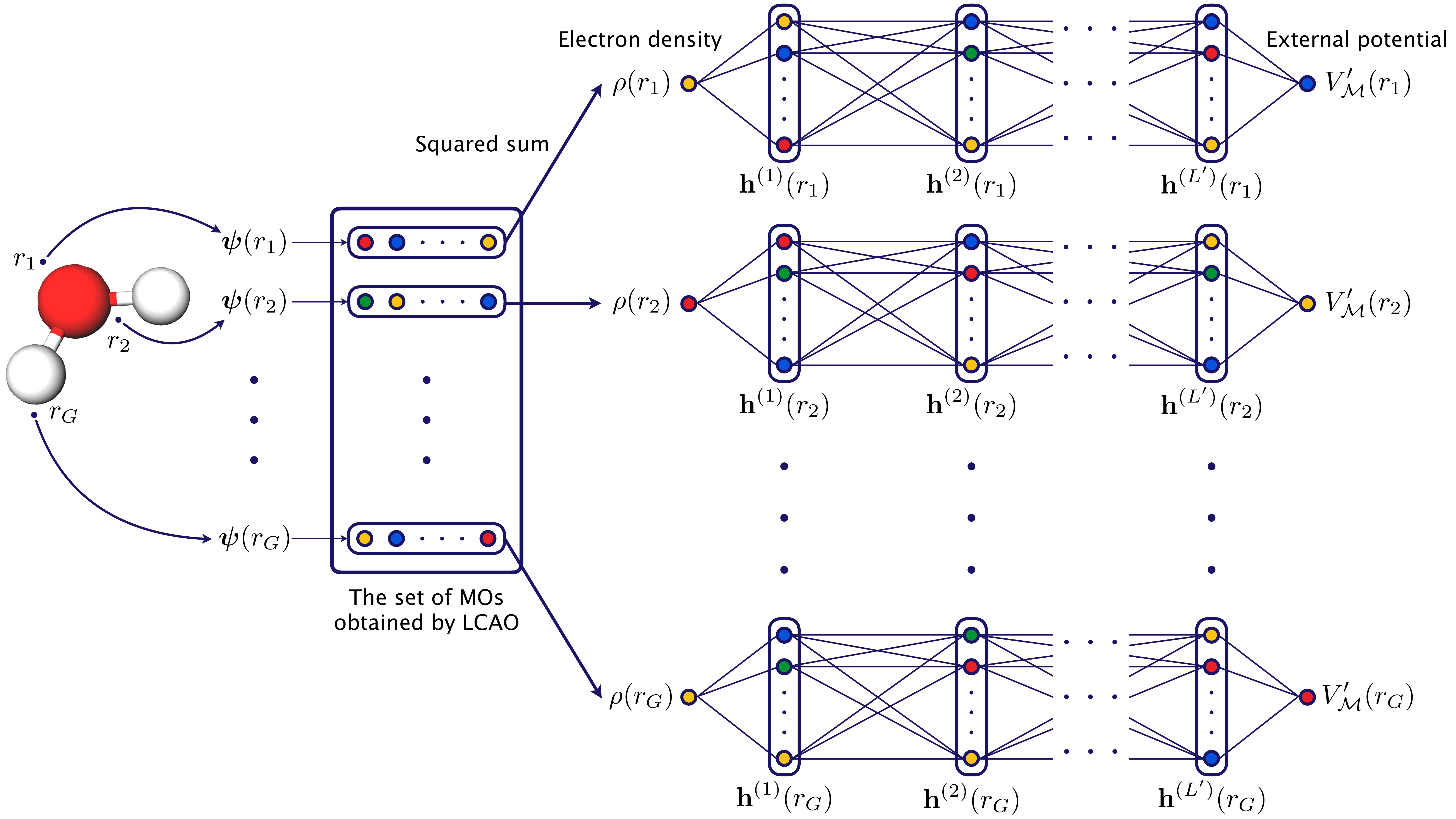}
\caption
{\label{fig_HKDNN}
Architecture of our DNN-based Hohenberg--Kohn (HK) map $\mathcal{HK}_{\text{DNN}}$.
}
\end{center}
\end{figure}

Using a similar feedforward architecture,
the DNN-based HK map $\mathcal{HK}_{\text{DNN}}$ is given by:
\begin{eqnarray}
&& \mathbf h^{(1)}(r_i) = \mathbf w_{\rho} \rho(r_i) + b_{\rho},
\\
&& \mathbf h^{(\ell+1)}(r_i) = \text{ReLU} (
\mathbf W^{(\ell)}_{\text{HK}} \mathbf h^{(\ell)}(r_i)
+ \mathbf b^{(\ell)}_{\text{HK}} ),
\\
&& V'_{\mathcal M}(r_i) = \mathbf w^{\top}_V \mathbf h^{(L')}(r_i) + b_V.
\end{eqnarray}
In this $\mathcal{HK}_{\text{DNN}}$, each hidden layer is
$\mathbf h \in \mathbb R^{N'}$, where $N'$ is a hyperparameter.
Figure~\ref{fig_HKDNN} illustrates the architecture of this DNN-based HK map.

{\bf Optimization.}
Using back-propagation and an SGD (in practice,
this study used the Adam optimizer~\cite{kingma2014adam}),
we minimize the loss function $\mathcal L_E$ in the main text;
in other words, we update the set of learning parameters
$\Theta_E = \{ \{ \zeta_n \}_{n=1}^N, \{ \mathbf c_n \}_{n=1}^N,
\{ \mathbf W^{(\ell)}_E \}_{\ell=1}^L, \{ \mathbf b^{(\ell)}_E \}_{\ell=1}^L,
\mathbf w_E, b_E \}$ as follows:
\begin{eqnarray}
\Theta_E \leftarrow \Theta_E - \alpha \frac{1}{B} \sum_{k=1}^B
\frac {\partial \mathcal L_{E_{\mathcal M_k}}} {\partial \Theta_E},
\end{eqnarray}
where $\mathcal L_{E_{\mathcal M_k}}$ is the energy loss value of
the $k$th molecule $\mathcal M$ in the training dataset,
$\alpha$ is the learning rate, and $B$ is the batch size.
Additionally, we also minimize the loss function
$\mathcal L_V = \sum_{i=1}^{G} ||V_{\mathcal M}(r_i) - V'_{\mathcal M}(r_i)||^2$,
i.e., we update the set of learning parameters
$\Theta_V = \{ \{ \zeta_n \}_{n=1}^N, \{ \mathbf c_n \}_{n=1}^N,
\mathbf w_{\rho}, b_{\rho}, \{ \mathbf W^{(\ell)}_{\text{HK}} \}_{\ell=1}^{L'},
\{ \mathbf b^{(\ell)}_{\text{HK}} \}_{\ell=1}^{L'},
\mathbf w_V, b_V \}$ as follows:
\begin{eqnarray}
\Theta_V \leftarrow \Theta_V - \alpha \frac{1}{B} \sum_{k=1}^B
\frac {\partial \mathcal L_{V_{\mathcal M_k}}} {\partial \Theta_V},
\end{eqnarray}
where $\mathcal L_{V_{\mathcal M_k}}$ is the potential loss value of
the $k$th molecule $\mathcal M$ in the training dataset.
We emphasize that QDF updates $\Theta_E$ and $\Theta_V$ ``alternately'' and
note that the learning parameters in the LCAO,
i.e., $\{ \zeta_n \}_{n=1}^N$ and $\{ \mathbf c_n \}_{n=1}^N$,
are ``shared'' in $\Theta_E$ and $\Theta_V$.

{\bf Normalization.}
The LCAO considers the normalization for the coefficients in Eq.~(6) in the main text.
This can be implemented by updating the coefficients in the iterative SGD as follows:
\begin{eqnarray}
\mathbf c'_n \leftarrow \frac{\mathbf c'_n}{|\mathbf c'_n|}.
\end{eqnarray}
Note that $\mathbf c'_n \in \mathbb R^N $ is not the $n$th row vector
but the $n$th column vector of matrix $\mathbf C$
described in Figure~4 in the main text.
Additionally, the normalization term
in Eq.~(7) in the main text is calculated as follows:
\begin{eqnarray}
Z(q_n, \zeta_n) =
\int |D_n^{(q_n-1)} e^{-\zeta_n D_n^2}|^2 dD =
\sqrt{\frac {(2q_n-3)!! \sqrt{\pi/2}} {2^{2(q_n-1)} \zeta_n^{(2q_n-1)/2}}}.
\end{eqnarray}
Note that because each Gaussian expansion $\zeta_n$
is a learning parameter of the model, $Z(q_n, \zeta_n)$ is recalculated
every time the model parameters are updated in SGD.

{\bf Hyperparameters.}
All model and optimization hyperparameters and their values used in this study
are listed in Table~\ref{hyperparameters}.

\begin{table}[t]
\begin{center}
\begin{tabular}{lr}
\hline \hline
Hyperparameter & Value \\
\hline
Sphere radius $s$ & 0.75 \AA \\
Grid interval $g$ & 0.3 \AA \\
\# of dimensions $N$ & 200 \\
\# of hidden layers in $\mathcal{F}_{\text{DNN}}$ & 3 or 6 \\
\# of hidden units in $\mathcal{HK}_{\text{DNN}}$ & 200 \\
\# of hidden layers in $\mathcal{HK}_{\text{DNN}}$ & 3 or 6 \\
Batch size & 4, 16 \\
Learning rate & 1e-4 or 5e-4 \\
Decay of learning rate & 0.5 \\
Step size of decay & 200 or 300 epochs \\
Iteration & 2000 or 3000 epochs \\
\hline \hline
\end{tabular}
\end{center}
\caption{
\label{hyperparameters}
List of all model and optimization hyperparameters and their values.}
\end{table}

\end{document}